\documentclass[11pt]{article}
\usepackage{ACL2023}

\usepackage{xurl}
\usepackage{hyperref}
\hypersetup{breaklinks=true}

\usepackage{times}
\usepackage{latexsym}

\usepackage[T1]{fontenc}

\usepackage[utf8]{inputenc}

\usepackage{microtype}

\usepackage{amsmath,amssymb,amsfonts}
\usepackage{graphicx}
\usepackage{textcomp}
\usepackage{xcolor}
\usepackage{booktabs}
\usepackage{multirow}
\usepackage{inconsolata}
\usepackage{graphicx}
\usepackage[noend]{algpseudocode} 

\usepackage{listings}
\usepackage{spverbatim}
\usepackage{xspace}

\usepackage[LGR,T1]{fontenc}

\newcommand{\unipy}{{\texttt{\sc UniPy}}\xspace}
\usepackage{enumitem}

\usepackage[normalem]{ulem}
\newcommand{\ensuretext}[1]{#1}
\newcommand{\nertcomment}[4]{\ensuretext{\textcolor{#3}{[\ensuretext{\textcolor{#3}{\ensuremath{^{\textsc{#1}}_{\textsc{#2}}}}} #4]}}}

\newcommand{\an}[1]{\nertcomment{A}{A}{magenta}{#1}}

\usepackage{tikz}
\usetikzlibrary{shapes.geometric}
\usetikzlibrary{patterns}
\usepackage{pgfplots}
\usetikzlibrary{backgrounds}
\usetikzlibrary{matrix,calc}
\usetikzlibrary{svg.path}
\usetikzlibrary{automata,positioning,decorations.text,topaths,arrows.meta,decorations.pathmorphing,quotes}
\usepackage{bbding}

\title{Automated Python Translation}

\author{Joshua Otten \\ George Mason University \\ \texttt{jotten4@gmu.edu} \And
Antonios Anastasopoulos \\ George Mason University \\ \texttt{antonis@gmu.edu} \And
Kevin Moran \\ University of Central Florida \\ \texttt{kpmoran@ucf.edu}}

\begin{document}
\maketitle
\begin{abstract}
Python is one of the most commonly used programming languages in industry and education.
Its English keywords and built-in functions/modules allow it to come close to pseudo-code in terms of its readability and ease of writing.
However, those who do not speak English may not experience these advantages.  In fact, they may even be hindered in their ability to understand Python code, as the English nature of its terms creates an additional layer of overhead.
To that end, we introduce the task of automatically translating Python's natural modality (keywords, error types, identifiers, etc.) into other human languages.  This presents a unique challenge, considering the abbreviated nature of these forms, as well as potential untranslatability of advanced mathematical/programming concepts across languages. 
We therefore create an automated pipeline to translate Python into other human languages, comparing strategies using machine translation and large language models.
We then use this pipeline to acquire translations from five common Python libraries (\texttt{pytorch}, \texttt{pandas}, \texttt{tensorflow}, \texttt{numpy}, and \texttt{random}) in seven languages, and do a quality test on a subset of these terms in French, Greek, and Bengali.
We hope this will provide a clearer path forward towards creating a universal Python, accessible to anyone regardless of nationality or language background.\footnote{Link to Github repository, containing code and results: \url{https://github.com/Joshua-Otten/AutomatedPythonTranslation}}

\end{abstract}

\section{Introduction}

Python is not only growing to be one of the most well-known programming languages by emerging developers today, but perhaps becoming one of the most popular as well~\cite{pythonpopularity}. 
It is used extensively in industry, and especially in education, where teachers can leverage the English nature of its keywords and built-in functions and modules to allow students to understand code on a more simplified level. 
In this way Python mimics programmable pseudocode, allowing programmers to specify in near-language terms what the code is meant to do instead of memorizing superfluous lists of terminology and acronyms, worrying themselves with low-level details.
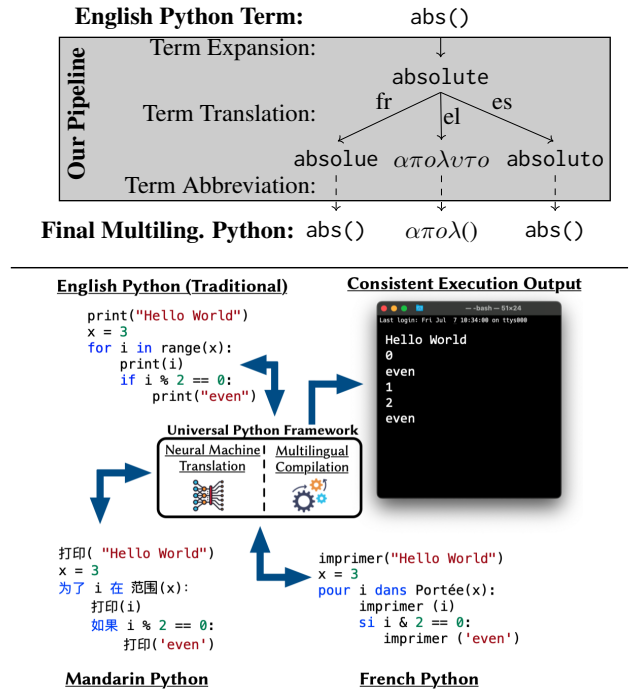
\begin{figure}[t]
    \centering
    \begin{tabular}{c}
    \resizebox{\columnwidth}{!}{
    \begin{tikzpicture}[shorten >=1pt,node distance=2cm,on grid,auto]
\draw[black, fill = black, fill opacity = 0.2, semithick] 
        (-4,0.6) rectangle (4.7,3.2);
\node[anchor=west,rotate=90] (text) at (-3.7,.8) {\textbf{Our Pipeline}};
\node[anchor=east] (a) at (0,3.5) {\textbf{English Python Term:}};
\node[anchor=west] (a1) at ($(a.east) + (40pt,0pt)$) {\texttt{abs()}};
\node[anchor=east] (b) at (.2,3) {Term Expansion:};
\node[anchor=north] (b1) at ($(a1.south) + (0pt,-10pt)$) {\texttt{absolute }};
\node[anchor=east] (c) at (0.2,2) {Term Translation:};
\node[anchor=west] (c1) at ($(b1.south) + (-70pt,-30pt)$) {\texttt{absolue}};
\node[anchor=west] (c2) at ($(b1.south) + (-25pt,-30pt)$) {$\alpha\pi o\lambda \upsilon\tau o$};
\node[anchor=west] (c3) at (c2.east) {\texttt{absoluto}};
\node[anchor=east] (d) at (0.2,.85) {Term Abbreviation:};
\node[anchor=north] (d1) at ($(c1.south) + (0pt,-17pt)$) {\texttt{abs()}};
\node[anchor=north] (d2) at ($(c2.south) + (0pt,-17pt)$) {$\alpha\pi o\lambda$()};
\node[anchor=north] (d3) at ($(c3.south) + (0pt,-17pt)$) {\texttt{abs()}};
\node[anchor=east] (e) at (-0.15,0.1) {\textbf{Final Multiling. Python:}};
\draw[ultra thin,->] (a1.south) -- (b1.north);
\draw[ultra thin,->] (b1.south) --  node[above,yshift=0pt,xshift=-2pt] {fr} (c1.north);
\draw[ultra thin,->] (b1.south) -- node[above,yshift=-7pt,xshift=5pt] {el} (c2.north);
\draw[ultra thin,->] (b1.south) -- node[above,yshift=0pt,xshift=2pt] {es} (c3.north);
\draw[ultra thin,dashed,->] (c1.south) -- (d1.north);
\draw[ultra thin,dashed,->] (c2.south) -- (d2.north);
\draw[ultra thin,dashed,->] (c3.south) -- (d3.north);
\end{tikzpicture}}\\
\midrule
\includegraphics[width=.9\columnwidth]{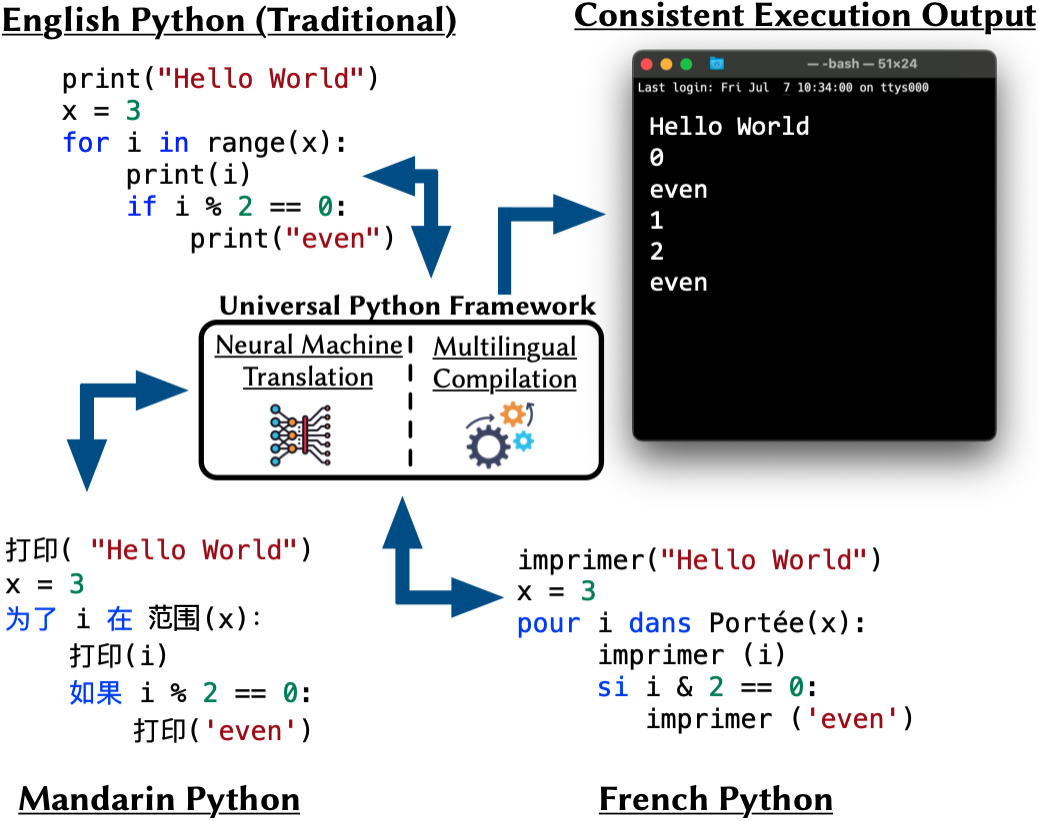}
\end{tabular}
    \caption{\text{Illustration} of our pipeline (top) with an example of the function \texttt{abs()} (absolute), which first expands English terms and then translates them into various languages, optionally abbreviating them. These translations could then be fed into the Universal Python Framework (bottom) from \citet{ottentowards}.}\vspace{-1.5em}
    \label{fig:pipeline}
\end{figure}

\iffalse
\begin{figure}[t]
    \centering
    \includegraphics[width=.95\columnwidth]{unipl-overview.png}
    \caption{Concept illustration for the Universal Python Framework. This framework translates natural language elements of the Python programming language into different natural languages other than English~\cite{ottentowards}.  Here we aim to automate the process.}\vspace{-1.5em}
    \label{fig:langselect}
\end{figure}
\fi

However, while English-speakers enjoy these advantages, non-English-speakers may struggle to learn or memorize these seemingly-strange terms. 
Several studies show that the speed with which a student learns to program correlates with their understanding of the human language it utilizes~\cite{scratchstudy}. %,piech2020human}.  
Additionally, \citet{piech2020human} found that many users write comments/commit messages in GitHub in their native languages, suggesting that people desire to code in non-English languages.
\textit{If we want computer science and technology to be accessible to all people of all cultures, then why is our code English-oriented?}  
The NLP community has seen a growing shift in attempting to broaden perspectives to other cultures.  
But when all of the tools are built with English programming languages, what does that say about the underlying system?
A first step to rectifying this is making Python multilingual.

We are aware of two recent approaches that attempt to tackle this problem: \texttt{CodeInternational}~\cite{piech2020human}, which automatically translates comments, identifiers, and (optionally) string literals in Python/Java code, and \unipy{}~\cite{ottentowards}, which deterministically translates the natural modality of the code itself, leaving comments/identifiers untouched for consistency. 
This translation is reversible, and allows the execution of non-English code.
Figure \ref{fig:pipeline} outlines the \unipy{} framework, and Table \ref{tab:compare} details similarities/differences between the two approaches. 
While \texttt{CodeInternational} translates programmer-specified aspects of the code, it does nothing to translate the natural modality of the code itself.
It may be useful for tracing and collaboration between people of different languages, but unlike \unipy{}, it does not allow someone to write and execute code from scratch that is meaningful to them in their own language.  
Therefore, we feel that \unipy{} presents a superior approach for our endeavor of making Python universal.

However, \unipy's translations were, in large part, constructed by hand using native-speaking annotators.
While this process worked well for a small prototype, Python has an ever-growing set of libraries, with the most recent tally at 137,000 according to Coding Ninjas~\cite{coding-ninjas}.  
When this number of packages is taken with the multitude of human languages for which \unipy could be expanded, hand-annotating Python's terms is simply not scalable.
If a truly universal Python is to be established, it would be very helpful to find a way of automating the translation process. 

Automating this task is more difficult than it may seem at first glance due to two major factors (i) \textit{conventions for writing identifiers}, and (ii) \textit{ambiguity in mapping technical terms to human languages}.  Python's terms largely consist of concatenated abbreviations (i.e., \texttt{snake\_case} identifiers with abbreviations such as \texttt{nan\_to\_num()}), which can confuse automated translators.  Additionally, mathematical or context-specific terms may not map to a single word in another language, complicating the translation process.

Therefore, our work provides three contributions toward reaching the ideal implementation of a ``universal Python:''
\begin{itemize}[noitemsep,nolistsep,leftmargin=*]
    \item First, we introduce the task of \textit{automatically} translating Python's terms into other human languages.
    \item Then, we create a pipeline for this process after comparing state-of-the-art models/methods.
    \item Finally, we use this pipeline to expand the current base of Python translations from \unipy{}, extending them to five additional common libraries (\texttt{pytorch}, \texttt{tensorflow}, \texttt{pandas}, \texttt{numpy}, and \texttt{random}) in seven human languages (Spanish, French, Greek, Hindi, Bengali, Mandarin, Arabic), and evaluate its effectiveness on a subset of these terms.
\end{itemize}

We also consider the effectiveness of fine-tuning an LLM to translate Python code, and provide analysis and discussion of this and our pipeline results.

\begin{table}[t]
    \centering
    \small
    \begin{tabular}{lccc}
    \toprule
        \textbf{Feature/translates} & \texttt{CodeInter} & \textbf{\unipy{}} & \textbf{} \\
    \midrule
    \multicolumn{3}{l}{\textbf{Code component}}\\
       \ comments/identifiers
         & \textcolor[rgb]{0.0, 0.75, 0.0}\checkmark & \textcolor[rgb]{0.8, 0.0, 0.0}\XSolidBrush \\
        \ natural modality & \textcolor[rgb]{0.8, 0.0, 0.0}\XSolidBrush & \textcolor[rgb]{0.0, 0.75, 0.0}\checkmark \\
        \multicolumn{3}{l}{\textbf{Capabilities}}\\
        \ code execution & \textcolor[rgb]{0.0, 0.75, 0.0}{\checkmark} & \textcolor[rgb]{0.0, 0.75, 0.0}\checkmark \\
        \ deterministic & \textcolor[rgb]{0.8, 0.0, 0.0}\XSolidBrush & \textcolor[rgb]{0.0, 0.75, 0.0}\checkmark \\
        \ Right-To-Left code & \textcolor[rgb]{0.8, 0.0, 0.0}\XSolidBrush & \textcolor[rgb]{0.0, 0.75, 0.0}\checkmark \\
    \bottomrule
    \end{tabular}
    \caption{Compares/contrasts \texttt{CodeInternational} with \unipy{}.  The functionality of these two projects is fundamentally different; \texttt{CodeInternational} translates/transliterates comments/identifiers/strings, while \unipy{} translates the natural modality of the code.}\vspace{-1.5em}
    \label{tab:compare}
\end{table}

\iffalse
\begin{figure*}[t]
    \centering
    \includegraphics[width=1.0\textwidth]{pipeline-figure.png}
    \caption{Concept Illustration for a pipeline that attempts to automatically translate Python libraries into other human languages.\an{This is (a) too big; (b) too vague: why are some things blue, red, or green, oval or square? Make this as minimal as possible. I can try making a mock.]}}
    \label{fig:pipeline}
\end{figure*}
\fi

\section{Automated Python Translation}
\label{sec:pipeline-creation}
Unlike most translation pipelines, converting Python's keywords and built-in functions/modules to a second language presents a unique challenge. 
While nearly all of the terms represent a word/phrase in English, most are abbreviated and/or concatenated with other words or abbreviations.  Automatic translators can take these forms literally; 
for instance, Google Translate interprets the \texttt{abs} from Python's "absolute value" function as short for "abdominal muscles."  

Furthermore, certain questions arise regarding grammatical function when translating to a language altogether different in structure.  For instance, the term "print" in the English language is inherently ambiguous--it could be a verb as in "to print" something, or a noun meaning "the print."  
If it is a verb, there is ambiguity regarding its nature.  Is it a command, an indicative action, or a suggestion?  Who is the one printing?  Since the English language is not morphologically rich, such information is not marked and the interpretation is left to the reader.   
However, when translating into a language such as Spanish, where verbal grammar is built in to the word itself, one must understand the nature of what is being translated, knowing the part of speech, tense, mood, person, etc.  Human translators of Python might disagree on these implicit grammatical assumptions, which complicates the overall process and could lead to inconsistency in term translation across libraries, etc.

Our intuition is that a pipeline for this task (outlined in Figure~\ref{fig:pipeline}) should solve several problems:
\begin{enumerate}
    \item \textbf{Python Term Expansion}, that is, un-abbreviate and un-concatenate if they are not proper English words, allowing translators to better understand the meaning behind what is being translated.
    \item \textbf{Python Term Translation} of expansions to a target language in a consistent manner.
    \item \textbf{Python Term Abbreviation}, where, possibly, one may abbreviate and/or concatenate translations.  We implement an (optional) rudimentory abbreviation scheme for our pipeline.
    For details, see Appendix \ref{sec:appendix-abbreviation}.
\end{enumerate}

Each of these tasks should be as automated and human-free as possible, to better allow a large number of Python libraries to be translated into just as many languages. 
Below, we first explore methods for term expansion and then term translation, before putting together a pipeline comprised of our best approaches to evaluate on new Python libraries for three languages.

\section{Python Term Expansion}

Many Python terms are abbreviations and/or concatenations of existing English words. 
Therefore, in order to process them so they can be more accurately translated, we attempt to ``expand'' these into unabreviated, standard English phrases.  As in Figure~\ref{fig:pipeline}, we provide original Python terms, and the model should output the proper English words or phrases that they represent.  For instance, \texttt{abs} would be expanded to ``absolute'' (value), and \texttt{delattr} would become ``delete attribute.''

\subsection{Experimental Settings}
\label{subsec:expansion-settings}

\paragraph{Data}
We evaluate expansion of the 222 unique terms from the Python standard library, testing model outputs against the hand-expanded forms from \citet{ottentowards}.

\paragraph{Models}

We use zero, one, and few-shot prompting with GPT-4 Turbo\footnote{gpt-4-0125-preview}~\cite{gpt-4-turbo} 
to expand Python standard library terms.  We also evaluate with a ``naive'' baseline that does not modify any terms; this simulates what accuracies would result if no expansion had been done in the first place. 

\paragraph{Prompts}
\label{subsec:expansion-prompts}
We try four different prompting strategies--zero-shot, zero-shot with motivation, one-shot, and 5-shot--in order to determine which prompting strategy will provide the most reliable output.  Following are examples of our prompts.  
\begin{itemize}[noitemsep,nolistsep,leftmargin=*]
    \item \texttt{0-shot:} 
    We use the following instruction:
    ``Please expand (i.e. split and unabbreviate) these Python terms into the word or phrase that they are intended to represent.  If no abbreviation or splitting into separate words is necessary, then the expanded form will be the same as the original term.  Do not provide any other response; simply list each term (each on a separate line) followed by => and its corresponding expansion (as in `[term] => [expansion]').  Here are the terms, separated by commas:  ''
    \item \texttt{0+Motive:} We augment the previous prompt to provide a motivation for the task.
    ``I am trying to translate Python's key terms into other languages, so that people can code in their native language.  However, I first need to know the expanded form of the abbreviations.  Please ... [same as above]'' 
    
    \item \texttt{1-shot:} Same as above, but now we add one example in the prompt: 
    ``[same as above] ... 
    
    For example: \texttt{abs} => \texttt{absolute value}.  Please expand these terms: ''
    \item \texttt{5-shot:} 
    As above, but instead we provide five examples, as below:
    ``[as above]... 
   
    For example: \texttt{abs} => \texttt{absolute value}, \texttt{memoryview} => \texttt{memory view}, \texttt{pow} => \texttt{power}, \texttt{print} => \texttt{print}, \texttt{SyntaxError} => \texttt{Syntax Error}.  Please expand these terms: ''
\end{itemize}
We compare this with a ``\texttt{naive}'' baseline that leaves the input as is (performing no expansion).

\subsection{Metrics}
\label{subsec:expansion-metrics}
We evaluate the above strategies using exact match accuracy and chrF scores (using \texttt{sacrebleu} package).  
Exact match provides a solid base with which to roughly judge how often a model was able to translate/expand the term \textit{exactly} how the humans did.  However, we also need to account for slight differences in spelling and grammar, or if a translation was partially correct.  Therefore, we also use chrF~\cite{popovic2015chrf}, since that judges character-level similarity for $n$-grams rather than word similarity, which we need given the short nature of Python terms.

\subsection{Results}
Our results for expansion are outlined in Table~\ref{tab:expansion-results}.  All approaches outperform the naive baseline. 
We obtain our best score for exact match accuracy (93.2\%) in the \texttt{5-shot} setting.  
ChrF scores tend to be quite close to each other, and we feel that the differences here are negligible.
Therefore, it seems reasonable to conclude that \texttt{5-shot} is currently the best setting for ChatGPT.

The fact that few-shot seems better than a 1-shot prompt suggests something significant -- %that in line with our intuition, 
this expansion task is novel.  The idea that LLMs need little instruction-tuning to perform a task they are pre-trained for is supported by findings from \citet{min2022rethinking}, \citet{xie2022explanation}, and \citet{ratner2023parallel}.  Therefore, these results suggest that Python term expansion may be a completely new task, and one that could be improved by showing examples of this during pre-training.

Regardless, accuracies this high are very promising that GPT expansion will be helpful in our translation pipeline.  Since the baseline scores are also relatively high (84.9\% for chrF), these results do not necessarily speak to GPT-4's overall ability; they do however demonstrate that for our task of Python term expansion, using an instruction-tuned LLM such as ChatGPT may be a reasonable approach.

\begin{table}[t]
    \centering
    \begin{tabular}{rcc}
    \toprule
        \textbf{Prompt} & \textbf{Accuracy} & \textbf{chrF} \\
    \midrule
        naive baseline & 46.9 & 84.9 \\
        \midrule
        \texttt{0-shot\ \ } & 89.6 & \textbf{96.1} \\
        \texttt{0+Motive} & 92.3 & 95.4 \\
        \texttt{1-shot\ \ }  & 91.0 & 95.4 \\
        \texttt{5-shot\ \ }  &  \textbf{93.2} & \textbf{95.7} \\
    \bottomrule
    \end{tabular}
    \caption{Expansion accuracy of Python's standard library using ChatGPT-4 Turbo on four prompts, showing both raw and chrF scores. Base represents the baseline of original (unmodified) Python terms.  In this case, \texttt{5-shot} (5-shot) clearly performs with the highest accuracy, suggesting that more context may be beneficial.}\vspace{-1em}
    \label{tab:expansion-results}
\end{table}

\section{Python Term Translation}
\label{sec:translation}

In order to determine the best translation strategy for the second stage of our pipeline, we test and evaluate three primary models with different strategies:  Google Translate, ChatGPT-4 Turbo, and Llama2.
We evaluate using the same metrics as in the expansion experiments (\S\ref{subsec:expansion-metrics}): exact match accuracy and chrF score.

\subsection{Experimental Settings}

\paragraph{Data}
For each system/method, we evaluate translations of the expanded form of the 222 unique Python standard library keywords and built-in functions/modules, acquired from \citet{ottentowards}.  Our references include eight languages in all: Spanish, French, Greek, Mandarin, Hindi, Begnali, Sorani Kurdish, and Arabic.\footnote{While the Arabic translations were vetted to be used in prototypes, several translations (around 17) were marked as not confident before we conducted these experiments; we mitigate this by leaving out these terms from our prompts.}

\paragraph{Models}

For our models, we opted to use Google Translate (as a baseline machine translation system), GPT-4 Turbo, and Llama2 with 70 billion parameters.  We also provide translation scores from ChatGPT-3 Turbo and Davinci models~\citet{gpt-3.5} in Appendix~\ref{sec:appendixA}, showing the extent to which version might affect in ChatGPT's abilities. 

\paragraph{Prompts}

We craft prompts for the two LLMs using similar strategies to \S\ref{subsec:expansion-prompts}.  As before, we have \texttt{0-shot}, \texttt{1-shot}, and \texttt{5-shot} prompts. We also consider that it might help a model to see preexisting translations in another language (to better understand the task), so we include an additional prompt (referred to here as \texttt{all-other}) with an entire set of the 222 terms.  For consistency, we chose Spanish as this reference for all languages except itself, in which case we use French examples. All complete prompts are listed in Appendix~\ref{sec:appendix-complete-prompts}.

As a translation model and not a LLM, these types of prompting strategies are not applicable to Google Translate.  However, we can emulate this methodology by providing three levels of context in the source sentence (the input to be translated), which we refer to as \texttt{no-cntxt}, \texttt{def}, and \texttt{expl}.
Our first level of context (\texttt{no-cntxt}) is similar to a zero-shot prompt--we simply translate the expanded forms of the terms.  
The second level (\texttt{def}) provides additional context by providing the term and a Python definition for it, and the final contextual level (\texttt{expl}) provides a sentence explanation of the term in question, followed by a colon and the term itself, as in the following examples:
\begin{itemize}[noitemsep,nolistsep,leftmargin=*]
    \item \texttt{no-cntxt}: ``print''
    \item \texttt{def}: ``print: Prints to the standard output device''
    \item \texttt{expl}:  ``In Python, to use the expression that prints to the standard output device, write: print.''
\end{itemize}

\begin{table*}[t]
    \centering
    \small
    \begin{tabular}{@{}c@{}c@{ \ }r@{ \ \ }l@{ \ \  }r@{ \ }l@{ \ \  }r@{ \ }l@{ \ \ }r@{ \ \  }l@{ \ \ }r@{ }l@{ \ }r@{ }l@{ \ }r@{ }l@{ \ }r@{ }l@{}}
    \toprule
    \multirow{2}{*}{\textbf{Model}} &  \multirow{2}{*}{\textbf{Prompt}} &  \multicolumn{2}{c}{\textbf{Spanish}}  & \multicolumn{2}{@{}c}{\textbf{French}} & \multicolumn{2}{@{}c}{\textbf{Greek}} &
    \multicolumn{2}{@{}c}{\textbf{Mandarin}} & \multicolumn{2}{@{}c}{\textbf{Hindi}} & \multicolumn{2}{@{}c}{\textbf{Bengali}} & \multicolumn{2}{@{}c}{\textbf{Arabic}} & \multicolumn{2}{@{}c}{\textbf{Kurdish}}  \\
   & & Raw & chrF & Raw & chrF & Raw & chrF & Raw & chrF & Raw & chrF & Raw & chrF & Raw & chrF & Raw & chrF  \\
    \midrule
     \multirow{5}{*}{\textsc{GPT-4}} &  \texttt{0-shot}  & \textbf{71.6}  &  83.4 &  \textbf{62.6} & 78.0 & 43.2 & 64.3 & 54.0 & 69.9 & 13.5 & 23.8 & 34.2 & 56.6 & 26.6 & 52.1 & 31.1 & 50.6 \\
                         &  \texttt{0+Motive}  & 68.9 & \textit{85.2} & 60.8 & 77.5 &  41.0 & 63.9 & \textbf{56.3} & \textit{71.0} & 28.8 & 49.8 & 31.1 & 55.1 & 27.5 & \textit{54.2} & \textbf{35.6} & \textit{54.4}   \\
                         &  \texttt{1-shot} & 71.2 & 83.8 & 61.7 & 78.3 & \textbf{45.1} & \textit{69.1} & 50.5  & 67.5 & 29.7 & 51.0 & 32.0 & 55.0 & \textbf{29.7} & 54.1 & 30.6 & 52.0 \\
                         &  \texttt{5-shot}  & 67.6  & 81.7 & 61.3 & \textit{78.7} & 44.1 & 66.6 & 27.5 & 37.9 & \textbf{35.6} & \textit{55.1} & \textbf{36.9} & \textit{57.6} & 13.5 & 27.9 & 34.2 & 54.3  \\
                         & \texttt{all-other} & 70.3 & 83.6 & 61.2 & 78.3 & 44.1 & 66.3 & 51.8 & 69.2 & 29.3 & 50.6 & 28.4 & 56.2 & 25.7 & 51.6 & 28.4 & 50.7 \\

    \midrule
    \multirow{5}{*}{Llama2} &  \texttt{0-shot}  &  \textbf{66.2} & \textit{81.5} &  49.6 & 68.9 & 12.6  & 34.8 & 27.0 & 42.2 & 5.9 & 19.1 & \textbf{5.0} & \textit{17.3} & 12.6 & 30.8 & 1.8 & 9.2  \\
                             &  \texttt{0+Motive}  &  59.0 & 76.5 & 49.6 & 69.7 & 13.5  & \textit{35.8} & 25.2 & 41.1 & 6.3 & 20.9 & 4.5 & 16.3 & 9.5 & 28.1 & 1.8 & 9.8  \\
                             &  \texttt{1-shot}  &  59.0  & 75.2 & 46.4 & 66.9 & 13.1  & 34.2 & 26.6 & 44.4 & 7.2 & 21.4 & \textbf{5.0} & 15.9 & \textbf{14.0} & 29.9 & 2.3 & 10.3  \\
                             &  \texttt{5-shot}  & 57.2 & 76.7  & 48.2 & 68.4 & \textbf{16.2} & 35.3  & \textbf{27.9} & \textit{45.7} & \textbf{10.8} & \textit{27.1} & 2.3 & 13.4 & 12.2 & \textit{31.0} & \textbf{2.7} & \textit{11.4} \\
                             & \texttt{all-other} & 59.0 & 78.8 & \textbf{50.0} & \textit{71.2} & 15.3 & 34.4 & 24.3 & 41.4 & 6.8 & 21.6 & 3.2 & 13.1 & 11.3 & 30.3 & 1.8 & 10.0 \\
                        
    \midrule
    \multirow{3}{*}{Google} &  \texttt{no-cntxt}  &  \textbf{73.4}  & \textit{82.4} & \textbf{84.2}  & \textit{88.8} & \textbf{45.5} & \textit{64.3} & \textbf{80.6} & \textit{86.4} & \textbf{39.2} & \textit{56.6} & \textbf{67.6} & \textit{80.5} & \textbf{61.3} & \textit{73.5} & \textbf{98.2} & \textit{98.6}  \\
                         &  \texttt{def}  & 55.9 & 73.1 & 43.2 & 61.1 & 32.9 & 46.1 & 25.2 & 29.3 & 19.4 & 32.1 & 18.5 & 27.5 & 23.0 & 51.4 & 23.9 & 27.4  \\
                         &  \texttt{expl}  &  0.5  & 66.7 & 16.2  & 39.1 & 27.0 & 37.1 & 1.4 & 3.2 & 30.2 & 45.9 & 32.4 & 47.3 & 23.4 & 47.0 & 23.4 & 32.3 \\

    \bottomrule
    \end{tabular}
    \caption{Python translation quality: exact match (raw) accuracy and chrF score, for each prompt (\texttt{0-shot}, \texttt{0+Motive}, etc.) or contextual level (\texttt{no-contxt}, \texttt{def}, \texttt{expl}), for all translation models.  The best raw scores per prompt are \textbf{bolded}, while best chrF scores are \textit{italicized}.  Google Translate without context is consistently the best among models across all languages, often by large margins. 
    }
    \label{tab:complete-results}
    \vspace{-1.5em}
\end{table*}

\subsection{Results}

Our translation results are in Table~\ref{tab:complete-results}.  
We evaluate using exact match accuracy (raw) and chrF scores.  Note that chrF scores rarely match corresponding raw accuracies perfectly.  

When a chrF score is a bit lower than the raw accuracy, we can assume that non-matching translations were simply wrong.  
On the other hand, a higher chrF score indicates that, even when translations were wrong, some portions were indeed correct.

Additionally, the best raw score per prompt is not always the same as the highest chrF score.  To determine which (if any) prompting strategy is optimal, we perform several ANOVA tests on these results (for tables, see App~\ref{sec:appendixANOVA}).  However, in all these cases, no prompting strategy can be said to be better than another at a 90\% level of confidence.

\paragraph{ChatGPT}

ChatGPT-4 demonstrates noteworthy ability in term translation, especially chrF.  
 
Table~\ref{tab:complete-results} shows relatively good scores for Spanish and French, nearly in the 70s range, and hovering around 80 for chrF.  With Greek and Mandarin the scores begin to drop, and then they continue down to as low as 29\% raw accuracy.  We found these differences in language scores to be statistically significant for both Raw and chrF accuracy at 99\% confidence, using single Factor ANOVA tests according to scores averaged across languages.\footnote{These results can be found in Appendix Table~\ref{tab:anova-turbo-lang}.}  
This analysis suggests that ChatGPT can be fairly trusted for high resource languages, which resonates with other multilingual LLM findings~\citet{hendy2023good}.  However, as the level of resources for a language drops, so does its accuracy.  As far as which of the five prompting strategies gives us the best performance, we used ANOVA testing for the five prompting ``alternatives'' (see Appendix Table~\ref{tab:anova-turbo-prompt}) and found that there are no statistically significant differences among them; our prompting strategies did not have enough of an effect on ChatGPT's performance to suggest that one is better than another.

\paragraph{Llama2}

Overall, Llama2 appears to perform even worse than ChatGPT; Table~\ref{tab:complete-results} shows that it is outperformed for almost every data point.  Though we again see statistical significance with respect to scores across languages (see App. Table~\ref{tab:anova-llama-lang}), our ANOVA tests revealed no significant differences based on prompting strategy.

\paragraph{Google Translate}

Google Translate without additional context (\texttt{no-cntxt}) yields the best results, outperforming ChatGPT in most of the cases.
 
It also generally appears to have a more even distribution with regard to scores across languages, although ANOVA tests reveal statistical significance in the scores across languages at 90\% confidence.  

Regarding the effect of context in translation, Table~\ref{tab:complete-results} shows a clear drop-off from \texttt{no-cntxt} to \texttt{def}, and then \texttt{def} to \texttt{expl}, where \texttt{expl} (containing the most context) is absolutely terrible.  Using ANOVA and the Method of Contrasts, we found the differences between the versions to be statistically significant with 99\% confidence, both for raw and chrF scores (see App. Table~\ref{tab:anova-goog}). 

Upon analysis of the translations, we find that in the majority of cases, \texttt{expl} did not even attempt to translate the Python term, instead translating everything but the terms in question.  This indicates that Google Translate was perhaps given too much context--the longer sentences reveal that the terms are indeed Pythonic and therefore were not translated, which is reasonable given that these are indeed typically not translated across languages.
In other words, since Python's key terms are historically only English, the translation model reasonably opted to keep them untranslated so that the initial context could be maintained.

\paragraph{Difficulties with Parsing and Formatting}

Responses from ChatGPT and Llama2 sometimes proved difficult to parse and/or format.  For instance, they often provided terms in mixed scripts, or simply restated the Python terms to translate.  Even after receiving formatting specifications, they would occasionally neglect to translate or expand a particular term, and omit it in the list of outputs.  In these cases we simply used ``-'' in place of a translation.  Interestingly, certain terms were more frequently omitted than others; for example, Llama2 often neglected to translate \texttt{as}.

\section{Downstream Pipeline Evaluation}

After evaluating various models' ability to expand and translate the Python standard library, we now try our best pipeline in the wild, on terms from five \textit{additional} Python libraries--\texttt{Pandas}, \texttt{Pytorch}, \texttt{TensorFlow}, \texttt{Numpy}, and \texttt{Random}--in seven languages (es, fr, el, hi, bn, ar, zh-cn). 
We manually extract terms from online documentation, and our resulting translation set comprises
6,119 terms.  
These can be made to work with \unipy{} simply by augmenting its dictionary lists with these terms.

We evaluate a subset of our outputs from these libraries in 

Greek, Bengali, and French.  We use native speakers to hand-annotate results (i.e. correct pipeline outputs).

Since abbreviating terms is optional in our pipeline, we only evaluate translations for these experiments. 

\subsection{Setup}

We combine our best techniques from the expansion and translation tasks to create an optimal pipeline.  
GPT-4 was able to expand Python terms with impressive accuracy, but due to cost constraints we use GPT-3.5 Turbo for this expansion\footnote{gpt-3.5-turbo-1106}~\cite{gpt-3.5}.
While not as impressive as GPT-4, GPT-3 Turbo can still perform with significant accuracy in the few-shot setting (for numbers on this, see Appendix \ref{sec:appendixA}), so we believe this will work as a first attempt.
As for term translation, we observe that \texttt{no-cntxt} of Google Translate had the best performance of all models.
Therefore, our pipeline has the following steps:
\begin{enumerate}[noitemsep,nolistsep]
    \item Expand the terms using ChatGPT-3 Turbo, prompting with a similar scheme as the 5-shot example in \S\ref{subsec:expansion-prompts}.
    \item Translate the processed form of the terms using Google Translate \texttt{no-cntxt} (no additional context).  We also do some minor post-processing (replacing spaces with underscores and removing determiners).
\end{enumerate}

\noindent We test this for Greek, Bengali and French, on four common Python libraries, \texttt{tensorflow}, \texttt{pandas}, \texttt{pytorch}, \texttt{random}, and \texttt{numpy},\footnote{Note that since these libraries are extensive, we only test with a subset of terms.} translating a total of 6,119 terms. 
We hand-annotate (correct) a subset of the outputs (407) by asking if each given translation could be considered reasonable, and if not, correcting to something that is.
Note that this process matches the envisioned scenario of language communities contributing to correct and solidify the automatically produced outputs.
We evaluate with raw and chrF scores.

\subsection{Results}

The scores for our pipeline translations (Table \ref{tab:pipeline-results}) 
vary significantly, but are overall quite good considering the novelty of this task.  
For instance, chrF scores remain above 60\% for all languages on average, and Bengali achieves over 90\% in over half the cases.
This demonstrates an important find: our pipeline can already do a fairly good job at initial translations of Python terms from scratch.  Without improvements, this method should significantly cut the time needed for manual annotation.
Also of note are the two differing ways Bengali-speakers express mathematical concepts: transliterated English terms, and Bengali words.  
While we attempted translation here for more thorough analysis, transliteration should be straightforward to implement as well.  Ideally, future work would create Python versions for both so that speakers could use their preferred expressions.

Some examples where translations failed include improper/unhelpful phrasing and words used in the wrong context.  For instance in French, certain phrases such as ``argument partition'' were translated as ``partition d'argument'' but corrected to ``argument partition.''  None of the important vocabulary changed, but the phrasing was improved.  For words in an improper context, ``less than or equal'' translated to ``moins ou égal'' but was corrected to ``inférieur ou égal.''  Both ``moins'' and ``inférieur'' have similar translations, but a different context here that affects the overall meaning.
We also include some examples from Hindi. 
While we did not have resources to do a comprehensive evaluation of this language, we analyzed some outputs and found many mistranslations.
There were phrases that translated with inappropriate context from the English side; for instance ``uniform'' translated to the clothing rather than the distribution, ``keys'' translated to the tool rather than for ``key/value pair,'' and ``character'' translated to the persona in a story/play, rather than an alphanumeric representation.
Since the ambiguity inherently arises from the English, we suspect that other languages may have this issue as well.

Finally, 
occasionally ChatGPT would expand inappropriately, such as \texttt{set\_tooltips} expanding to Spanish ``establecer consejos.''
Sometimes ChatGPT's expansions can be extremely long, as in \texttt{random.rand} expanding to "random data or random values generated with uniform distribution."  This, while not inaccurate, is far too long to be used as a Python term. This case would benefit from our abbreviation scheme.

\begin{table*}[t]
    \centering
    \small
    \begin{tabular}{@{}rc@{ }cc@{ }cc@{ }cc@{ }cc@{ }cc@{ }c@{}}
    \toprule
           & \multicolumn{2}{c}{\textbf{\textsc{PyTorch} \%}} & \multicolumn{2}{c}{\textbf{\textsc{TensorFlow} \%}} & \multicolumn{2}{c}{\textbf{\textsc{Pandas}} \%} & \multicolumn{2}{c}{\textbf{\textsc{Random} \%}} & \multicolumn{2}{c}{\textbf{\textsc{NumPy} \%}} & \multicolumn{2}{c}{\textbf{Total \%}} \\
        \# terms & \multicolumn{2}{c}{80} & \multicolumn{2}{c}{80} & \multicolumn{2}{c}{80} & \multicolumn{2}{c}{22} & \multicolumn{2}{c}{145} & \multicolumn{2}{c}{407} \\
        metric & raw & chrF & raw & chrF & raw & chrF & raw & chrF & raw & chrF & raw & chrF \\
    \midrule
        French & 50.0 & 75.2 & 41.3 & 71.2 & 61.3 & 75.5 & 31.8 & 65.8 & 52.4 & 71.6 & \textbf{50.1} & \textbf{72.7} \\
        Greek & 32.5 & 63.7 & 38.8 & 64.3 & 33.8 & 59.9 & 27.3 & 51.2 & 46.2 & 66.1 & \textbf{38.6} & \textbf{63.2} \\
        Bengali & 98.8 & 99.7 & 82.5 & 90.9 & 96.3 & 97.7 & 72.7 & 73.5 & 53.8 & 67.8 & \textbf{82.1} & \textbf{90.8} \\
    \bottomrule
    \end{tabular}
    \caption{Results from the Pipeline experiment, translating 407 terms from five Python libraries into different languages.  The Total scores represent accuracies over a combined set of terms from the packages.}\vspace{-1em}
    \label{tab:pipeline-results}
\end{table*}

\section{Code-Block Translation Model}

In addition to our pipeline experiments, we consider the effectiveness of finetuning an LLM to translate entire blocks of Python code. 
We extract code samples from \texttt{codeparrot/github-code} and translate into four languages (Spanish, French, Greek, and Hindi) using \unipy. 
To ensure that all appearing terms are supported by either \unipy or our own translations, we filter by import statements.
We also filter prompts by a character length of 500 for efficiency.
Our resulting training set comprises 32,528 examples, where translations can be in either the English $\rightarrow$ non-English or non-English $\rightarrow$ English direction. We LoRA finetune the \texttt{Llama-2-7b-chat-hf} model for 15 epochs, and test on an additional set of 13,165 code examples, evaluating with BLEU~\cite{papineni2002bleu} and chrF score (both with \texttt{sacrebleu}). 
Results are in Table~\ref{tab:model-scores} in Appendix~\ref{sec:appendixFinetuning}.
Overall, we receive positive scores, demonstrating that an LLM can successfully translate Python code blocks.
However, the model over-generated in many cases, writing additional translated code, etc.
Given these clear imperfections and non-perfect scores, we cannot expect translations to be reliably executable.
However, this method may be useful for code accessibility (e.g. multi-lingual documentation), or perhaps extracting translations of terms that may then be included in the \unipy tables.

\section{Discussion}
\label{discussion}

All in all, our initial automated translation pipeline requires successful ability to perform two tasks: expansion of the original Python terms, and translation into a target language.  Fortunately, we find relatively high accuracies for both, using GPT-4's expansions under a few-shot prompt, and Google Translate \texttt{no-cntxt} for the translations.  

Asking ChatGPT to expand Python terms was met with positive results.
It appears that expanding Python terms may not be present in ChatGPT's training data (especially GPT-3, see Appendix \ref{sec:appendixA}), suggesting that our task is completely new.  
We would likely find even better results by including more examples in the prompts, or if ChatGPT were pre-trained for expansion in the future.
Google Translate (without context) achieved the highest translation accuracy and consistency across languages, demonstrating its superiority for this task. 
This may be partly due to issues with the pre-trained generative models failing to adequately follow instructions, paired with inadequate overall performance on lower-resourced languages such as Sorani Kurdish.  
We test this pipeline in the wild for the \texttt{pytorch}, \texttt{pandas}, \texttt{tensorflow}, \texttt{numpy}, and \texttt{random} libraries, obtaining scores mostly over 50\%, in some cases extremely high.  Thus, we can expect decent initial translations from our pipeline, especially for higher-resourced languages.  Overall, our pipeline provides a fast method of translating terms. %to other libraries.
We also try fine-tuning Llama2 to translate Python code blocks, and are met with positive results. 
However, these are not currently sufficient for our task of creating executable code.

In the practical setting, our pipeline could already be used to obtain preliminary translations for the rest of Python's multitude of libraries into other human languages.  
For languages with good performance, prototype versions could be developed and used out of the box, without necessarily requiring immediate annotation.  
Then, these initial translations could be made open-source and updated by native-speaking annotators all over the world for higher-quality versions.  
Once we have translations, all we need to do to integrate this into \unipy{} is update the mapping tables.

Future work should find or create even better methods for Python expansion and translation, to further alleviate the burden currently placed on annotators. 
One experiment might improve GPT prompting to include translated code segments, providing more context to better translate Python terms.
Additionally, while our pipeline includes an initial abbreviation scheme, it would be helpful to find language-specific methods of abbreviation to make terms more meaningful to programmers.

\section{Related Work}
While a Universal Python is still early in development, several instances of programming languages were developed for users of particular linguistic backgrounds.  For instance, Scratch and Blockly (used in the educational space) support certain non-English languages, and `KuMir'\footnote{\url{https://web.archive.org/web/20160112180533/http://lpm.org.ru/kumir2/}} and `Glossa'\footnote{\url{https://web.archive.org/web/20160112180533/http://lpm.org.ru/kumir2/}} are Pascal-based programming languages using key terms in Russian and Greek, respectively~\citet{codingforeveryone}.  
\citet{piech2020human} analyzed the extent of multilinguality on GitHub, and created a tool, ``\texttt{CodeInternational},'' to automatically translate identifiers defined in a Java or Python codebase (such as function names), comments, and optionally, string literals, to other languages using Google Translate.  While this approach can certainly be helpful, it does not translate the modality of the code itself, falling short of creating a ``universal'' Python.
\citet{ottentowards} began the process of manually translating Python's standard library into eight other human languages.  
We use these translations as references in our experiments.

\section{Conclusion}

Python translation is a necessary task, and a pipeline is essential for any large-scale translation efforts. 
We present the first-ever pipeline to do this and obtain reasonable results.
This paper introduces the task of automatically translating Python terms, building a pipeline consisting of three main steps:  expansion,
translation,  
and abbreviation. 

We use our best pipeline to translate four additional Python libraries, contributing over 6,000 new terms to the current base of Python translations in seven languages.  We perform a quality test on 407 of these translations for Greek, French, and Bengali, obtaining positive initial results with room for improvement.  
Although automated translation of Python is nowhere near perfect, we can begin the process of translating libraries for high-resourced languages and expect positive initial results. 
This is an important step toward universal programming, where everyone from any culture can code in their native language.

\section*{Limitations}

It is worth noting that human annotators may disagree as to what constitutes a reasonable translation; this may be a factor in certain model scores, especially across languages where annotators change.  For our work, we were only able to have one annotator per library.

\section*{Ethics Statement}
Using ChatGPT and Llama2's outputs may carry with it certain privacy concerns over where the training data for these LLMs came from, and how it was used in text generation, which is out of scope for this work.

\section*{Acknowledgements}
This work was generously supported by the Presidential Scholarship awarded by the George Mason University Graduate Division. It was also partially supported by NSF Award IIS-2327143.  
Computational resources for experiments were provided by the Office of Research Computing at George Mason University (URL: \url{https://orc.gmu.edu}). 
Also special thanks goes to the anonymous reviewers of this paper.

\bibliography{anthology,custom}
\bibliographystyle{acl_natbib}

\appendix

\section{Comparisons between LLM performance}
\label{sec:appendixA}
\begin{figure*}[t]
    \centering
    \includegraphics[width=.90\textwidth]{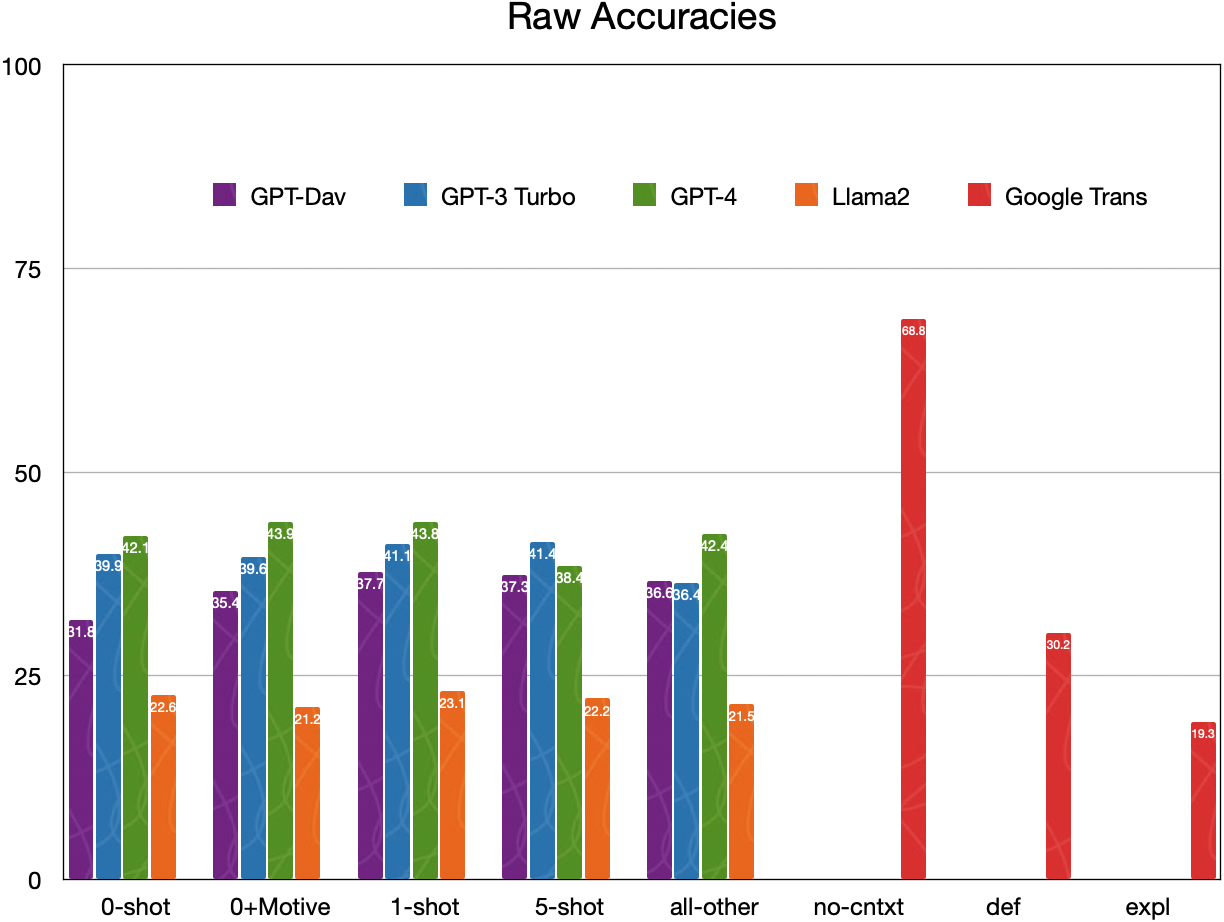}
    \caption{Plot of the raw accuracy percentage averaged over the languages, for each prompt/model.}
    \label{fig:prompt-graph-raw}
    \vspace{1em}
\end{figure*}

\begin{figure*}[t]
    \centering
    \includegraphics[width=1.0\textwidth]{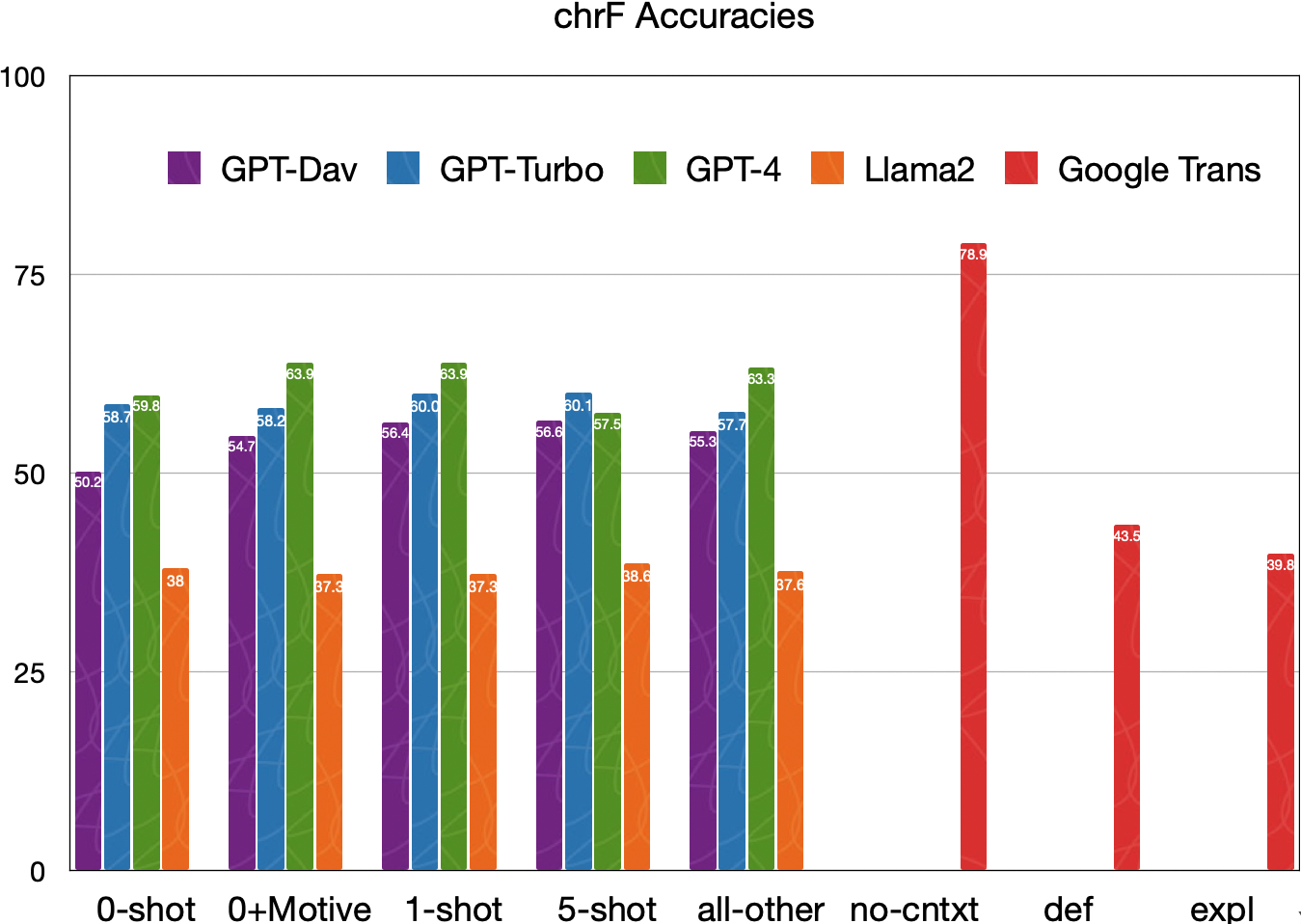}
    \caption{Plot of the chrF accuracy percentage averaged over the languages, for each prompt/model.}
    \label{fig:prompt-graph-chrf}
\end{figure*}

Table \ref{tab:gpt3-translation} provides translation accuracies for GPT-3 Turbo and GPT-3 Davinci, and Figures \ref{fig:prompt-graph-raw} and \ref{fig:prompt-graph-chrf} show comparisons between some of the LLM performance for each prompt, averaged across all languages.  It is worth noting that ChatGPT and Llama2's results appear somewhat worse in the graphs than they are in reality; this is due to the lower-resource languages deflating the overall averages (even though some languages, such as Spanish and French, achieved relatively high accuracy). 

\begin{figure*}[t]
    \centering
    \includegraphics[width=.90\textwidth]{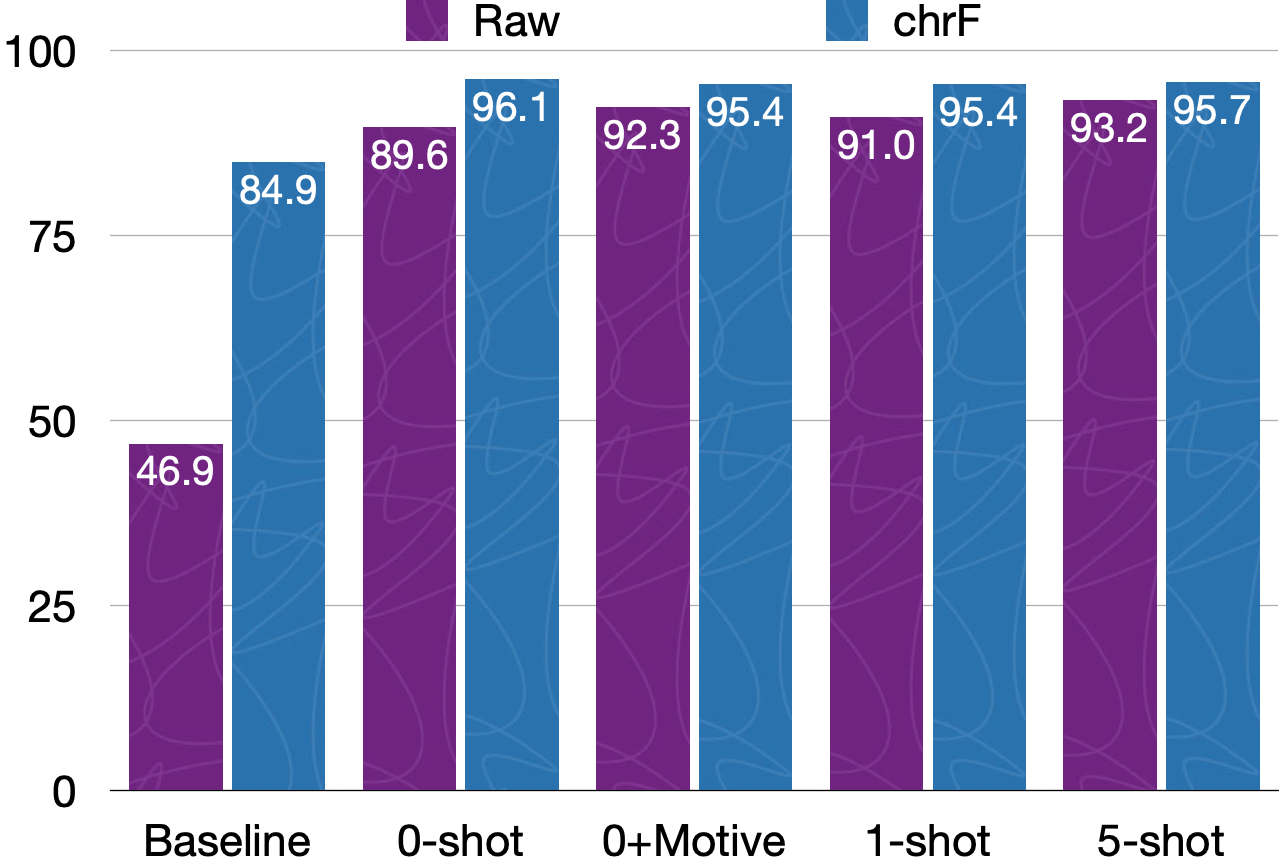}
    \caption{Plot of GPT-4's expansion accuracy percentage (Raw and chrF) for each prompt/model.}
    \label{fig:expansion-graph}
\end{figure*}

For expansion accuracies, Figure \ref{fig:expansion-graph} provides a comparison of the results for each prompt.  Note that terms were only expanded in English.

We additionally include expansion accuracies of GPT-3 Turbo in Table \ref{tab:expansion-results-turbo}, since we use it in our pipeline.

\begin{table*}[t]
    \centering
    \small
    \begin{tabular}{@{}c@{}c@{ \ }r@{ \ \ }l@{ \ \  }r@{ \ }l@{ \ \  }r@{ \ }l@{ \ \ }r@{ \ \  }l@{ \ \ }r@{ }l@{ \ }r@{ }l@{ \ }r@{ }l@{ \ }r@{ }l@{}}
    \toprule
    \multirow{2}{*}{\textbf{Model}} &  \multirow{2}{*}{\textbf{Prompt}} &  \multicolumn{2}{c}{\textbf{Spanish}}  & \multicolumn{2}{@{}c}{\textbf{French}} & \multicolumn{2}{@{}c}{\textbf{Greek}} &
    \multicolumn{2}{@{}c}{\textbf{Mandarin}} & \multicolumn{2}{@{}c}{\textbf{Hindi}} & \multicolumn{2}{@{}c}{\textbf{Bengali}} & \multicolumn{2}{@{}c}{\textbf{Arabic}} & \multicolumn{2}{@{}c}{\textbf{Kurdish}}  \\
   & & Raw & chrF & Raw & chrF & Raw & chrF & Raw & chrF & Raw & chrF & Raw & chrF & Raw & chrF & Raw & chrF  \\
    \midrule
    \multirow{5}{*}{\textsc{Turbo}} &  \texttt{0-shot}  & 70.3 & 83.9	& \textbf{64.0} & 78.3	& 41.4	& 64.1 & 52.7 & 68.1 & 	27.5 & 46.7 &	27.5 & 47.7 &	25.7 & 51.8 &	10.4 & 29.1 \\
                         &  \texttt{0+Motive}  & 71.2 & \textit{85.2} &	63.5 & 78.1 &	41.0 & 64.7 &	50.5 & 67.5 &	27.0 & 46.6 &	26.6 & 45.1 &	25.7 & 51.3 &	11.7 & 27.1  \\
                         &  \texttt{1-shot} & \textbf{73.0} & 84.7 &	57.7 & 76.5 &	\textbf{44.1} & 66.3 &	\textbf{54.5} & \textit{69.1} &	25.2 & 48.3 &	\textbf{32.0} & \textit{50.4} &	\textbf{27.9}	& \textit{52.7} & \textbf{14.4} & \textit{32.1} \\
                         &  \texttt{5-shot}  & 72.5	& 82.0 & 64.4 & \textit{79.6} &	43.7 & 66.3 & 	50.5 & 67.8 &	\textbf{28.8} & \textit{50.5} &	31.5 & 50.3 &	26.6 & 52.4 &	13.5 & 31.6  \\
                         & \texttt{all-other} & 55.4 & 80.7 &	60.4 & 77.4	& 44.1 & \textit{67.4} &	47.3 & 68.8 &	26.6	& 47.9 & 25.7 &  48.9 &	23.4	& 49.0 & 8.1 & 21.8 \\
    \midrule
     \multirow{5}{*}{\textsc{Davi}} &  \texttt{0-shot}  & 66.2   &  81.5 &  53.6 & 75.3 & 36.0 & 57.2 & 38.7 & 53.3 & 26.1 & 48.5 & 7.7 & 18.1 & 22.5 & 49.9 & 3.6 & 17.8\\
                         &  \texttt{0+Motive}  & 64.9 & \textit{82.2} & 55.4 & 75.1 & 36.5 & 59.3 & 46.0 & 61.1 & 22.1 & 43.0 & \textbf{28.4} & \textit{46.3} & 22.1 & \textit{50.1} & 8.1 & 20.4   \\
                         &  \texttt{1-shot} & 64.0 & 79.0 & 58.6 & 76.6 & \textbf{39.6} & \textit{62.5} & \textbf{53.6}  & 66.3 & \textbf{27.5} & \textit{49.6} & 23.0 & 41.8 & \textbf{25.2} & 48.8 & 9.9 & 26.2 \\
                         &  \texttt{5-shot}  & \textbf{68.0}  & 82.0 & 56.3 & 74.9 & 35.4 & 60.0 & 47.3 & 61.8 & \textbf{27.5} & 48.4 & 25.7 & 46.2 & 24.3 & 50.0 & \textbf{14.4} & \textit{29.5}  \\
                         & \texttt{all-other} & 65.8 & 81.4 & \textbf{62.2} & \textit{78.8} & 35.1 & 58.4 & 50.9 & \textit{66.0} & 27.0 & 48.5 & 23.9 & 43.1 & 22.1 & 48.5 & 5.9 & 17.5 \\
        \bottomrule
        \end{tabular}
    \caption{Translation results of the 222 standard library terms with GPT-3 Turbo and GPT-3 Davinci.  GPT-3 Turbo outperforms the Davinci model, but in general is not as good as GPT-4 Turbo.}
    \label{tab:gpt3-translation}
\end{table*}

\begin{table}[t]
    \centering
    \begin{tabular}{rcc}
    \toprule
        \textbf{Prompt} & \textbf{Accuracy} & \textbf{chrF} \\
    \midrule
        naive baseline & 46.9 & 84.9 \\
        \midrule
        \texttt{0-shot\ \ } & 70.7 & 82.8 \\
        \texttt{0+Motive} & 63.5 & 75.7 \\
        \texttt{1-shot\ \ }  & 53.2 & 73.0 \\
        \texttt{5-shot\ \ }  & \textbf{75.7} & \textbf{86.5} \\
    \bottomrule
    \end{tabular}
    \caption{Expansion accuracy of Python's standard library using ChatGPT-3 Turbo on four prompts, showing both raw and chrF scores. Base represents the baseline of original (unmodified) Python terms.  In this case, \texttt{5-shot} (5-shot) clearly performs with the highest accuracy, suggesting that more context may be beneficial.}
    \label{tab:expansion-results-turbo}
\end{table}

\section{Abbreviation Scheme}
\label{sec:appendix-abbreviation}

In an attempt to follow general Python conventions, we abbreviate according to syllable structure.  
For a given word, we first separate into syllables, where each consists of either a [vowel] or [set of consonants]+[vowel].  
Then, we abbreviate by keeping the first two syllables plus one additional consonant, discarding the rest of the word. 
If there is a collision, we iteratively add back letters until it is a unique term again.

If the term is multi-word (e.g. separated by underscores), we first eliminate unnecessary articles and coordinating conjunction words, and then abbreviate each individual word according to the above process.

We believe this works as an initial attempt to ensure that newly translated terms are short enough to be reasonably used in Python code for many languages. 
However, our current approach would not work with all languages (such as Mandarin) whose writing systems do not allow segmentation into syllabic structure. 
Additionally, languages may have differing conventions for abbreviation (or even none at all), in which case it will be necessary to develop more nuanced techniques to handle the abbreviation task.

\section{Finetuned Model Scores}
\label{sec:appendixFinetuning}
We provide a table of our finetuning results for translation of entire code blocks (Table~\ref{tab:model-scores}).
The scores demonstrate reasonable performance and suggest that this method has potential--however, considering the precise nature of computer programming, this is unlikely to be good enough for cases requiring execution of the translated code.

\begin{table*}[]
    \centering
    \begin{tabular}{ccccccccccc}
    \toprule
        Metric & En$\rightarrow$Es & Es$\rightarrow$En & En$\rightarrow$Fr & Fr$\rightarrow$En & En$\rightarrow$El & El$\rightarrow$En & En$\rightarrow$Hi & Hi$\rightarrow$En & Avg\\
    \midrule
        BLEU & 38.7 & 32.6 & 38.8 & 32.6 & 39.8 & 33.0 & 39.7 & 32.8 & \textbf{36.0} \\
        chrF & 66.3 & 59.4 & 66.3 & 59.7 & 62.9 & 59.3 & 64.1 & 59.6 & \textbf{62.2} \\
    \bottomrule
    \end{tabular}
    \caption{BLEU and chrF scores of Llama2 finetuned on translating code blocks. There appears to be little variation across languages; however interestingly we see that the English $\rightarrow$ non-English directionality performs better than the other way around.}
    \label{tab:model-scores}
\end{table*}

\section{ANOVA Tests}
\label{sec:appendixANOVA}
\begin{table}[t]
    \centering
    \resizebox{8cm}{2cm}{
    \begin{tabular}{p{0.6cm}cp{0.8cm}cp{0.6cm}cc}
    \toprule
         Variation & SS & df & MS & F & P-value & F crit \\
         \midrule
         SSA & 76.180	& 4 & 	19.045 &	0.063	& 0.992 &	2.113 \\
         SSE & 10521.084 &	35 &	300.602 \\
         Total & 10597.263 &	39 \\
        \midrule
         SSA & 266.589 & 4 & 66.647 & 0.280 & 0.889 & 2.113 \\
         SSE & 8319.351 & 35 & 237.696 \\
         Total & 8585.94 & 39 \\
    \bottomrule \\
    \end{tabular}}
    \caption{Results of ANOVA test on ChatGPT-4 Turbo's five prompting strategies, at a 90\% confidence interval.  The first set of rows indicates analysis for raw scores, while the second is chrF.  All values are rounded to 3 decimal places, for this and other ANOVA tables.  Since the P-value is greater than $\alpha = 0.1$, we can conclude that variation due to the prompts is not statistically significant at this level of confidence.}
    \label{tab:anova-gpt4-prompt}
\end{table}

\begin{table}[t]
    \centering
    \resizebox{8cm}{2cm}{
    \begin{tabular}{p{0.6cm}cp{0.8cm}cp{0.8cm}cc}
    \toprule
         Variation & SS & df & MS & F & P-value & F crit \\
         \midrule
         SSA & 9522.082	& 7 & 	1360.297 &	40.486	& 3.91E-14 &	3.258 \\
         SSE & 1075.181 &	32 &	33.599 \\
         Total & 10597.26 &	39 \\
        \midrule
         SSA & 6597.024 & 7 & 942.432 & 15.163 & 1.48E-08 & 3.258 \\
         SSE & 1988.916 & 32 & 62.154 \\
         Total & 8585.94 & 39 \\
    \bottomrule \\
    \end{tabular}}
    \caption{Results of ANOVA test on ChatGPT-4 Turbo's language scores, at a 99\% confidence interval. The first set of rows indicates analysis for the raw scores, while the second is for chrF.}
    \label{tab:anova-gpt4-language}
\end{table}

\begin{table}[t]
    \centering
    \resizebox{8cm}{2cm}{
    \begin{tabular}{p{0.6cm}cp{0.8cm}cp{0.6cm}cc}
    \toprule
         Variation & SS & df & MS & F & P-value & F crit \\
         \midrule
         SSA & 141.796	& 4 & 	35.449 &	0.102	& 0.981 &	2.091 \\
         SSE & 13945.827 &	40 &	348.646 \\
         Total & 14087.623 &	44\\
        \midrule
         SSA & 10.32 & 4 & 2.58 & 0.004 & 0.999 & 2.113 \\
         SSE & 21051.18 & 35 & 601.462 \\
         Total & 21061.5 & 39 \\
    \bottomrule \\
    \end{tabular}}
    \caption{Results of ANOVA test on ChatGPT-3 Turbo's five prompting strategies, at a 90\% confidence interval.  The first set of rows indicates analysis for raw scores, while the second is chrF.}
    \label{tab:anova-turbo-prompt}
\end{table}

\begin{table}[t]
    \centering
    \resizebox{8cm}{2cm}{
    \begin{tabular}{p{0.6cm}cccccc}
    \toprule
         Variation & SS & df & MS & F & P-value & F crit \\
         \midrule
         SSA & 13708.88	& 7 &	1958.412 &	172.650 &	1.44E-23 &	3.258 \\
         SSE & 362.985	32	& 11.343 \\
         Total & 14071.87	& 39 \\
         \midrule
         SSA & 20951.4 & 7 & 2993.057 & 869.917 & 1.22E-34 & 3.258 \\
         SSE & 110.1 & 32 & 3.441 \\
         Total & 21061.5 & 39 \\
        \bottomrule \\
    \end{tabular}}
    \caption{Results of ANOVA test on ChatGPT-3 Turbo's language scores, at a 99\% confidence interval.  The first set of rows indicates analysis for the raw scores, while the second is for chrF.}
    \label{tab:anova-turbo-lang}
\end{table}

% ChatGPT Davinci
\begin{table}[t]
    \centering
    \resizebox{8cm}{2cm}{
    \begin{tabular}{p{0.6cm}cp{0.8cm}cp{0.6cm}cc}
    \toprule
         Variation & SS & df & MS & F & P-value & F crit \\
         \midrule
         SSA & 179.471 & 4 & 44.868 & 0.113 &	0.977	& 2.113 \\
         SSE & 13857.014	& 35 &	395.915 \\
         Total & 14036.485 &	39 \\
        \midrule
         SSA & 215.066 & 4 & 53.767 & 0.138 & 0.967 & 2.113 \\
         SSE & 13674.02 & 35 & 390.686 \\
         Total & 13889.09 & 39 \\
    \bottomrule \\
    \end{tabular}}
    \caption{Results of ANOVA test on ChatGPT Davinci's five prompting strategies, at a 90\% confidence interval.  The first set of rows indicates analysis for raw scores, while the second is chrF.}
    \label{tab:anova-davinci-prompt}
\end{table}

\begin{table}[t]
    \centering
    \resizebox{8cm}{2cm}{
    \begin{tabular}{p{0.6cm}cccccc}
    \toprule
         Variation & SS & df & MS & F & P-value & F crit \\
         \midrule
         SSA & 13486.926	& 7	& 1926.704	& 112.189	& 1.107E-20	&3.258 \\
         SSE & 549.558	& 32 &	17.174 \\
         Total & 14036.485 &	39 \\
         \midrule
         SSA & 13035.34 & 7 & 1862.192 & 69.798 & 1.43E-17 & 3.258 \\
         SSE & 853.748 & 32 & 26.680 \\
         Total & 13889.09 & 39 \\
        \bottomrule \\
    \end{tabular}}
    \caption{Results of ANOVA test on ChatGPT Davinci's language scores, at a 99\% confidence interval.  The first set of rows indicates analysis for the raw scores, while the second is for chrF.}
    \label{tab:anova-davinci-lang}
\end{table}

\begin{table}[t]
    \centering
    \resizebox{8cm}{2cm}{
    \begin{tabular}{ccccccc}
    \toprule
         Variation & SS & df & MS & F & P-value & F crit \\
         \midrule
         SSA & 10.422	& 4 &	2.605 &	0.006 &	0.999 &	2.113 \\
         SSE & 16534.470 &	35 &	472.413 \\
         Total & 16544.892	& 39 \\
        \midrule
         SSA & 10.32 & 4 & 2.58 & 0.00429 & 0.999 & 2.113 \\
         SSE & 21051.18 & 35 & 601.462 \\
         Total & 21061.5 & 39 \\
        \bottomrule \\
    \end{tabular}}
    \caption{Results of ANOVA test on Llama2's five prompting strategies at 90\% confidence.  Raw on top, chrF on bottom.}
    \label{tab:anova-llama}
\end{table}

\begin{table}[t]
    \centering
    \resizebox{8cm}{2cm}{
    \begin{tabular}{p{0.6cm}cccccc}
    \toprule
         Variation & SS & df & MS & F & P-value & F crit \\
         \midrule
         SSA & 12613.351 	& 7 &	1801.907	& 817.728 &	5.848E-27 &	3.496 \\
         SSE & 52.885 &	24 &	2.204 \\
         Total & 12666.236	& 31 \\
         \midrule
         SSA & 20951.4 & 7 & 2993.053 & 869.917 & 1.22E-34 & 3.258 \\
         SSE & 110.1 & 32 & 3.441 \\
         Total & 21061.5 & 39 \\
        \bottomrule \\
    \end{tabular}}
    \caption{ANOVA test on Llama2's language scores, at a 99\% confidence interval.  The first set of rows indicates analysis for raw scores, while the second is for chrF.}
    \label{tab:anova-llama-lang}
\end{table}

\begin{table}[t]
    \centering
    \resizebox{8cm}{2cm}{
    \begin{tabular}{p{0.6cm}cccccc}
    \toprule
         Variation & SS & df & MS & F & P-value & F crit \\
         \midrule
         SSA & 10799.504	& 2	& 5399.752	& 22.559 &	5.889E-06 &	5.780 \\
         SSE & 5026.688	& 21 &	239.366 \\
         Total & 15826.192	& 23 \\
         \midrule
         SSA & 7444.426 & 2 & 3722.213 & 13.783 & 1.5E-04 & 5.780 \\
         SSE & 5671.164 & 21 & 270.055 \\
         Total & 13115.59 & 23 \\
        \bottomrule \\
    \end{tabular}}
    \caption{Results of ANOVA test on Google Translate's three contextual version strategies, at 99\% confidence.  Top rows are raw, bottom are chrF.}
    \label{tab:anova-goog}
\end{table}

\begin{table}[t]
    \centering
    \resizebox{8cm}{2cm}{
    \begin{tabular}{p{0.6cm}cccccc}
    \toprule
         Variation & SS & df & MS & F & P-value & F crit \\
         \midrule
         SSA & 922.228 & 7 & 131.747 & 0.141 & 0.993 & 2.128 \\
         SSE & 14903.964 & 16 & 931.498 \\
         Total & 15826.192 & 23 \\
         \midrule
         SSA & 2442.63 & 7 & 348.947 & 0.523 & 0.804 & 2.128 \\
         SSE & 10672.96 & 16 & 667.06 \\
         Total & 13115.59 & 23 \\
        \bottomrule \\
    \end{tabular}}
    \caption{ANOVA test on Google Translate's language scores, at a 90\% confidence interval.  The first set of rows indicates analysis for raw scores, while the second is for chrF.}
    \label{tab:anova-goog-lang}
\end{table}

In general for the LLMs, we find the variation of scores to be statistically insignificant across prompts, but significant across languages (see Tables \ref{tab:anova-gpt4-prompt}, \ref{tab:anova-gpt4-language}, \ref{tab:anova-turbo-prompt}, \ref{tab:anova-turbo-lang}, \ref{tab:anova-davinci-prompt}, \ref{tab:anova-davinci-lang}, \ref{tab:anova-llama}, \ref{tab:anova-llama-lang}).  This suggests that while our prompts did not have very much effect on the output, we can expect LLMs to perform much better on high resource languages than lower-resourced ones.

\paragraph{Google Translate}
At a 90\% confidence interval, neither the chrF or raw scores are statistically significant (see Tables \ref{tab:anova-goog} and \ref{tab:anova-goog-lang}). It should also be noted that at 99\%, language selection for chrF ceases to be relevant.

The raw accuracy for \texttt{def} and \texttt{expl} was significant at 95\% confidence intervals, but we cannot make this claim at the 99\% level.  On the other hand, the chrF scores for these were very similar, such that the differences were not found to be significant at even a 90\% confidence interval.

\section{Complete Prompts}
\label{sec:appendix-complete-prompts}
We list the complete prompts here for expansion and translation. 
\paragraph{Expansion}
\begin{itemize}
    \item \textbf{\texttt{0-shot}:} 
    "Please expand (i.e. split and unabbreviate) these Python terms into the word or phrase that they are intended to represent.  If no abbreviation or splitting into separate words is necessary, then the expanded form will be the same as the original term.  Do not provide any other response; simply list each term (each on a separate line) followed by => and its corresonding expansion (as in '[term] => [expansion]').  Here are the terms, separated by commas:  "
    \item \textbf{\texttt{0+Motive}:}
    "I am trying to translate Python's key terms into other languages, so that people can code in their native language.  However, I first need to know the expanded form of the abbreviations.  Please help me with this by expanding (i.e. splitting and unabbreviating) each of the following terms into the word or phrase that they are intended to represent. If no abbreviation or splitting into separate words is necessary, then the expanded form will be the same as the original term.  Do not provide any other response or translations; simply list each term (each on a separate line) followed by => and its corresonding expansion (as in '[term] => [expansion]').  Here are the terms, separated by commas:  "
    \item \textbf{\texttt{1-shot}:} 
    "I am trying to translate Python's key terms into other languages, so that people can code in their native language.  However, I first need to know the expanded form of the abbreviations.  Please help me with this by expanding (i.e. splitting and unabbreviating) each of the following terms into the word or phrase that they are intended to represent. If no abbreviation or splitting into separate words is necessary, then the expanded form will be the same as the original term.  Do not provide any other response or translations; simply list each term (each on a separate line) followed by => and its corresonding expansion (as in '[term] => [expansion]').  For example: abs => absolute value.  Please expand these terms: "
    \item \textbf{\texttt{5-shot}:} %\begin{spverbatim} %Do we want this here?  If so, we'd need to be consistent right?
    "I am trying to translate Python's key terms into other languages, so that people can code in their native language.  However, I first need to know the expanded form of the abbreviations.  Please help me with this by expanding (i.e. splitting and unabbreviating) each of the following terms into the word or phrase that they are intended to represent. If no abbreviation or splitting into separate words is necessary, then the expanded form will be the same as the original term.  Do not provide any other response or translations; simply list each term (each on a separate line) followed by => and its corresonding expansion (as in '[term] => [expansion]').  For example: abs => absolute value\\memoryview => memory view\\pow => power\\print => print\\SyntaxError => Syntax Error.  Please expand these terms: "
\end{itemize}
\paragraph{Translation}
\begin{itemize}
    \item \textbf{\texttt{0-shot}:}
    "Please translate the following terms into [language].  Do not provide any other response or translations; simply list each term (each on a separate line) followed by => and its corresonding translation (as in `[term] => [translation]'). Here are the terms, separated by commas:  "
    \item \textbf{\texttt{0+Motive}:}
    "I am trying to translate Python's key terms into other languages, so that people can code in [language].  Please help me with this by translating each of the following terms into [language]. Do not provide any other response or translations; simply list each term (each on a separate line) followed by => and its corresonding translation (as in `[term] => [translation]').  Here are the terms, separated by commas:  "
    \item \textbf{\texttt{1-shot}:}
    "I am trying to translate Python's key terms into [language], so that people can code in [language].  Do not provide any other response or translations; simply list each term (each on a separate line) followed by => and its corresonding translation (as in `[term] => [translation]').  For example: absolute value => [translation].  Please translate these terms into [language]: "
    \item \textbf{\texttt{5-shot}:} 
    "I am trying to translate Python's key terms into [language], so that people can code in [language].  Do not provide any other response or translations; simply list each term (each on a separate line) followed by => and its corresonding translation (as in `[term] => [translation]').  For example: absolute value => [translation]\\memory view => [translation]\\power => [translation]\\print => [translation]\\Syntax Error => [translation].  Please translate these terms into [language]: "
    \item \textbf{\texttt{all-other}:}
    "I am trying to translate Python's key terms into [language], so that people can code in [language].  For example, when translating Python to French, you have these translations: [set of English => French terms, separated by commas].  Please translate the following terms into [language].  Do not provide any other response or translations; simply list each term (each on a separate line) followed by => and its corresonding translation (as in `[term] => [translation]').  Here are the terms, separated by commas:  "
\end{itemize}

\end{document}